\documentclass{article}

\usepackage[preprint,nonatbib]{neurips_2021}

\usepackage[
    backend=bibtex,
    backref=true,
    hyperref=true,
    isbn=true,
    sorting=none,
]{biblatex}
\DefineBibliographyStrings{english}{%
  backrefpage = {cited on p.},
  backrefpages = {cited on p.},
}
\usepackage[utf8]{inputenc} 
\usepackage[T1]{fontenc}    
\usepackage{hyperref}       
\usepackage{url}            
\usepackage{booktabs}       
\usepackage{amsfonts}       
\usepackage{nicefrac}       
\usepackage{microtype}      
\usepackage{xcolor}         
\usepackage{float}          
\usepackage{multirow}

\usepackage{import}
\usepackage{xifthen}
\usepackage{pdfpages}
\usepackage{transparent}
\usepackage{xkeyval}

\usepackage{adjustbox}

\makeatletter
\newlength{\inkfig@width}
\define@key{inkfig}{width}{\setlength\inkfig@width{#1}}
\setkeys{inkfig}{width=\columnwidth}%

\newcommand{\inkfig}[2][]{%
  \setkeys{inkfig}{#1}
  \def\svgwidth{\inkfig@width}%
  \import{./figures/}{#2.pdf\string_tex}%
}
\makeatother

\title{AI Gone Astray: Technical Supplement} 

\author{%
  Janice Yang\thanks{These two authors contributed equally} \space $^\dagger{}$,~~~Ludvig Karstens\footnotemark[\value{footnote}] \space $^\dagger{}$,~ Casey Ross $^\ddag$, Adam Yala $^\dagger{}$\\[2em]
  $^\dagger{}$Department of Electrical Engineering and Computer Science\\
  Massachusetts Institute of Technology, Cambridge, MA, USA \\ [2em]
  $^\dagger{}$Jameel Clinic, Massachusetts Institute of Technology, Cambridge, MA, USA \\ [2em]
  $^\ddag$ STAT News, Boston, MA, USA \\ [2em]
  \texttt{\{janicey, ludvig\}@mit.edu} \\
  \texttt{casey.ross@statnews.com} \\
  \texttt{adamyala@csail.mit.edu} \\
}

\begin{document}

\maketitle
\setcounter{footnote}{0} 

\section{Introduction} 
This study is a technical supplement to ``AI gone astray: How subtle shifts in patient data send popular algorithms reeling, undermining patient safety.'' from STAT News\footnote{\url{www.statnews.com/2022/02/28/sepsis-hospital-algorithms-data-shift}}, which investigates the effect of time drift on clinically deployed machine learning models. We use MIMIC-IV, a publicly available dataset, to train models that replicate commercial approaches by Dascena and Epic to predict the onset of sepsis, a deadly and yet treatable condition. We observe some of these models degrade over time; most notably an RNN built on Epic features degrades from a 0.729 AUC to a 0.525 AUC over a decade, leading us to investigate technical and clinical drift as root causes of this performance drop. 

\section{Methods}
\begin{figure}[H]
    \centering
    \includegraphics[width=0.25\textwidth]{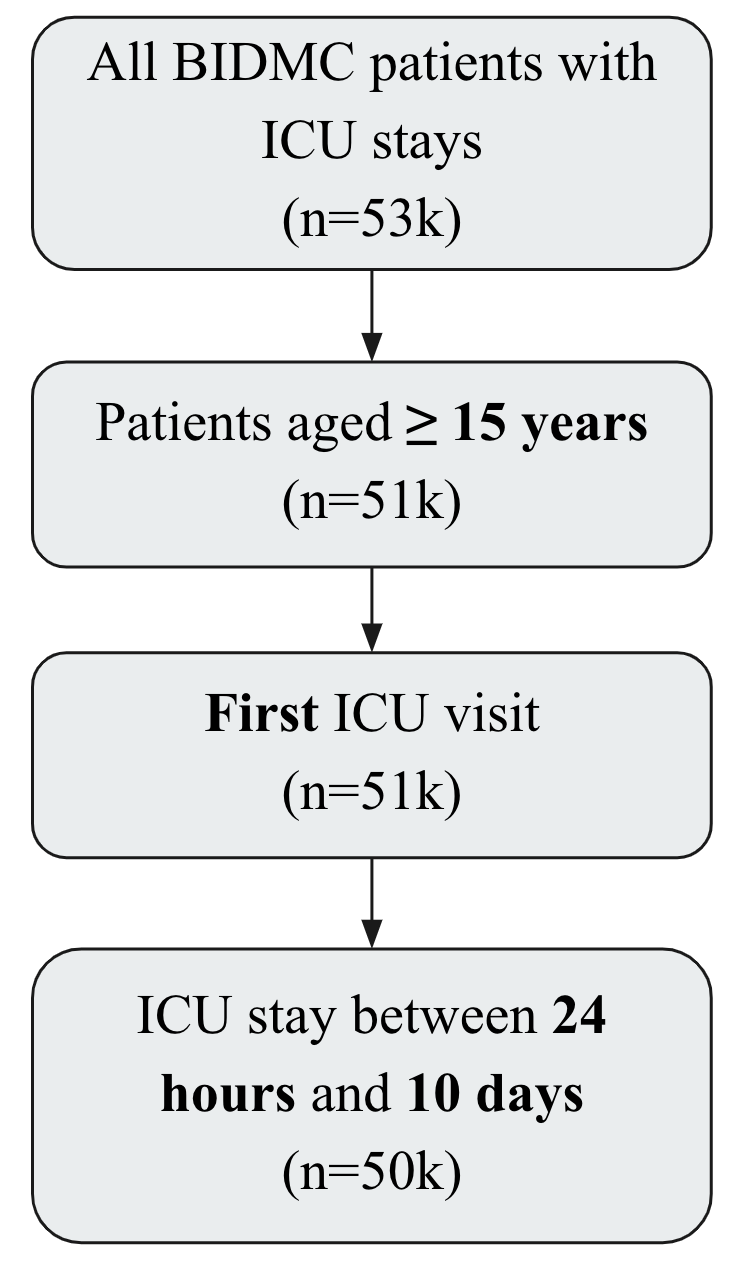}
    \caption{Cohort filtration process for MIMIC-IV}
    \label{fig:cohort_filter}
\end{figure}

\paragraph{Dataset}
We investigate time drift using the MIMIC-IV database \cite{johnson_alistair_mimic-iv_nodate}, which includes electronic health records of over 50,000 patients admitted to the intensive care units at the Beth Israel Deaconess Medical Center (BIDMC) between the years 2008-2019. We filter for patients over the age of 15, with an ICU stay between 24 hours and 10 days, and take each patient's first ICU stay (see \autoref{fig:cohort_filter}). After all filtering, we end up with 50k patients in our dataset. 

\paragraph{Sepsis Prediction Task}
Sepsis is a potentially life-threatening condition where a body's reaction to an infection triggers damage in organ systems. Following recent work \cite{moor_early_2019, moor_predicting_2021}, we predict the onset of Sepsis-3 within 6 hours given 24 hours of ICU data. Sepsis-3 defines sepsis onset as an increase in SOFA-score of >= 2 points within a window of 48 hours before and 24 hours after a Suspicion of Infection, which occurs when there are concomitant orders of antibiotics and microbiological samples taken. We follow the implementation of \citeauthor{moor_early_2019} \cite{moor_early_2019}, and use the SOFA value in the first hour of a patient’s ICU stay as the baseline score for later comparisons.





\paragraph{Cohort Construction}

\begin{figure}[H]
    \centering
    {\footnotesize
    \inkfig[width=0.8\textwidth]{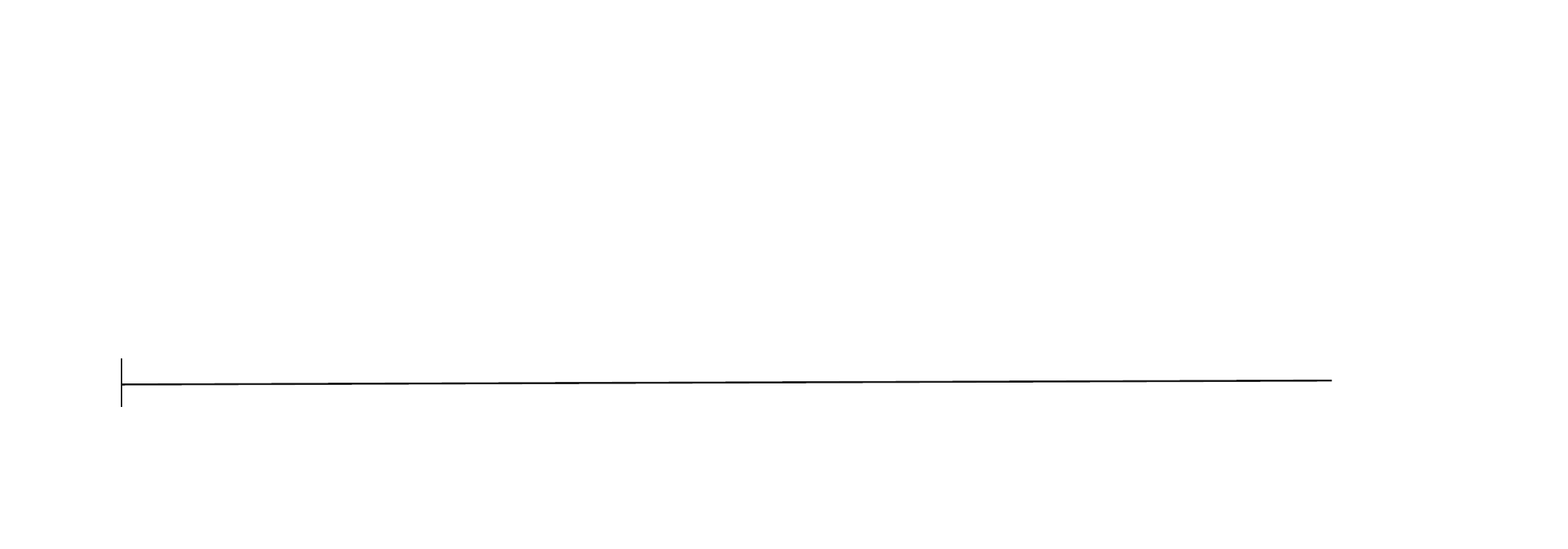}
    \inkfig[width=0.8\textwidth]{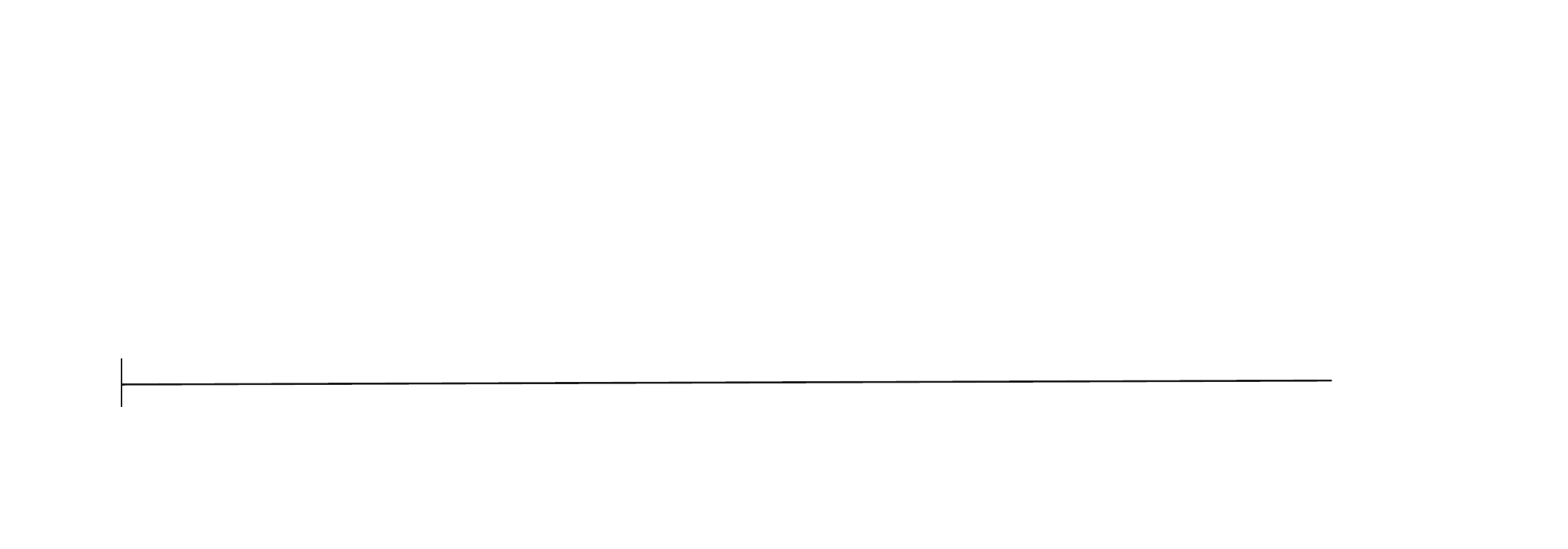}
    }
    \caption{Top: data construction process for a positive sepsis patient, taking their first sepsis onset time during their ICU stay. Bottom: data construction process for a control patient with no sepsis onset during their ICU stay, assigning a random "onset" time for data extraction purposes.}
    \label{fig:data_construction}
\end{figure}
See \autoref{fig:data_construction} for our cohort construction procedure. For patients with a positive sepsis label, the first instance of sepsis onset was taken to be the sepsis onset time. Following prior work by the Dascena group, \cite{barton_evaluation_2019}, we assign a random "onset" time for control patients with no sepsis onset during their stay. For both positive and control samples, we extract 24 hours of ICU data preceding a 6 hour gap period before their onset time. We eliminate all patients with an onset time before 6 hours. For patients who have less than 24 hours of data after extraction, we left pad all non-existent hours with zeros. 


\paragraph{Modeling Commercial Tools}
We model two commercial sepsis detection systems: Dascena's \cite{barton_evaluation_2019} and Epic Systems'. Their specific model implementations are not publicly available, so we use their known features to build our own models on MIMIC-IV and experiment with different model architectures. Dascena's sepsis models include a combination of six vital signs -- systolic blood pressure, diastolic blood pressure, heart rate, respiratory rate, peripheral oxygen saturation (SpO2), and temperature -- and patient age. Epic's sepsis models include more than 40 high level features, generally divided into 6 categories: Demographics, Vital Signs, Recent Lab Results, Chronic Illness Diagnoses, Medication Orders, and Active Drains, Airways or Wounds. 

These models leveraged two types of patient features: static demographic features, and time-varying features. For the former, one value per feature is included for each patient. For the latter, we follow prior work \autocite{nestor_feature_2019,moor_early_2019} and aggregate multiple irregularly-measured values into hourly buckets for each hour of a patient’s ICU stay. Vital sign measurements are aggregated by taking the average of all measurements in the hour, while events such as medication, infusions or drainages are aggregated by taking the sum across the hour. For each feature, we simply take their raw representation, as identified by each feature's unique item ID from the MIMIC-IV dataset. To account for data missingness in time-varying features, we implement simple imputation \cite{che_recurrent_2018}, which involves forward-filling each feature for each patient by hour, then concatenating binary flags for whether a feature was originally missing and another value for the time it was last recorded. A detailed list of the features we used and their Item ID's in MIMIC-IV can be found in Appendix A.


\subsection{Model Details}
To investigate the impact of model architecture on performance degradation, we experimented with both Logistic Regression models and Elman RNNs\cite{elman_finding_1990}. Reference our code repository for specific model implementation and training details. 

\begin{figure}[H]
\centering
\includegraphics[width=0.7\textwidth,clip]{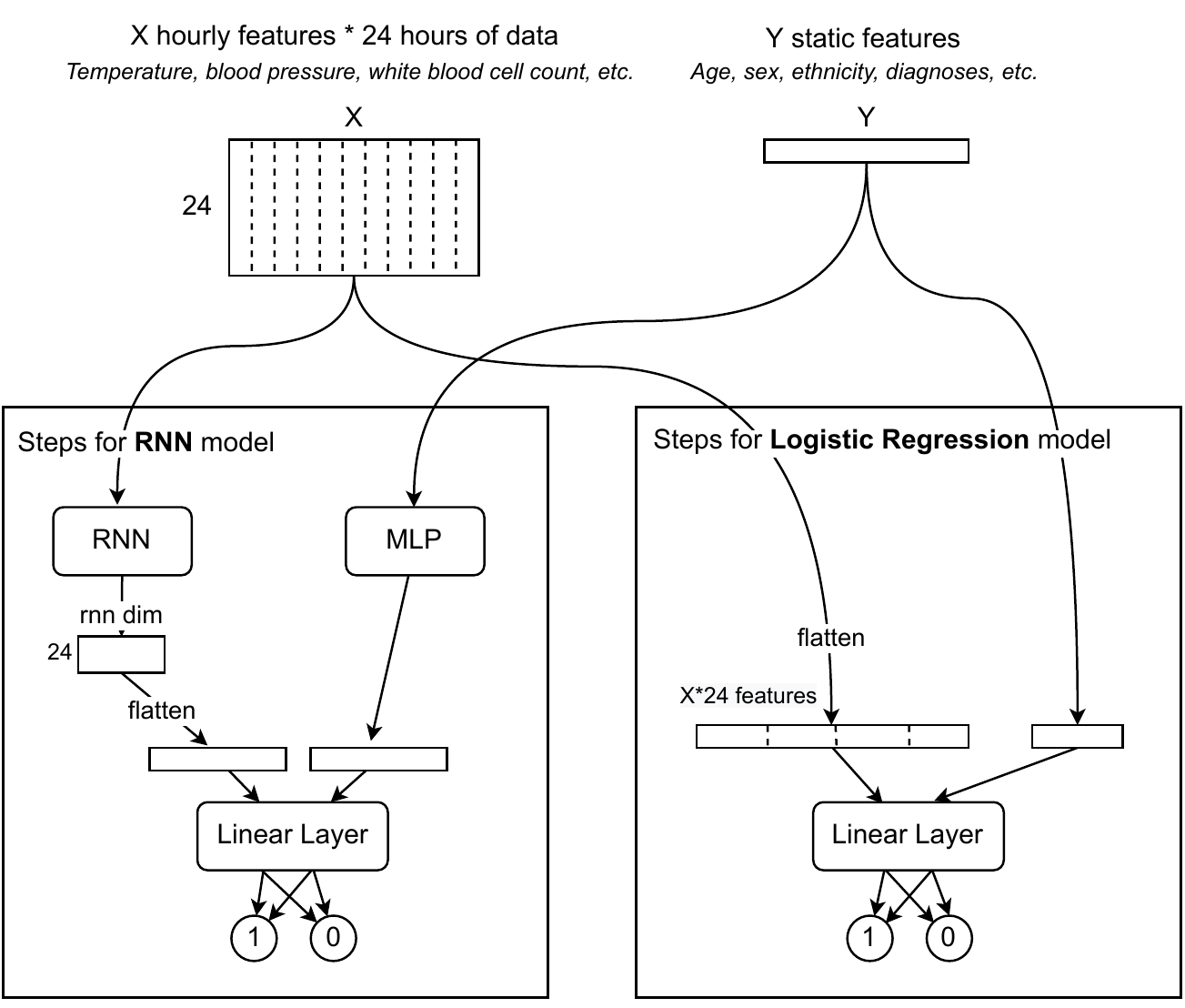}
\caption{Model architectures for Logistic Regression and RNN models}
\label{fig:models_inputs}
\end{figure}

\paragraph{Logistic Regression} Our Logistic Regression baseline is illustrated in \autoref{fig:models_inputs}. Following \citeauthor{nestor_feature_2019}\cite{nestor_feature_2019}'s work, we first flatten all hourly features, treating each measurement at each hour as a separate column. We then concatenate the result with the patient's static features to form the overall feature vector for the patient. In the context of Dascena features, static features include age, and hourly features include all vital sign measurements. In the context of Epic features, static features include demographic features and diagnoses codes, and hourly features include all vital sign measurements, medications, and lab measurements. 


\paragraph{RNN} To capture recurrent features, we implement an Elman RNN\cite{elman_finding_1990}. As depicted in \autoref{fig:models_inputs}, static features are first fed through a separate 4-layer MLP (of 32 neurons in each layer, preceded by batch-normalization and succeeded by ReLU activation), while hourly features are fed directly through the Elman RNN. Then, the output hidden representations from the MLP are concatenated with the output hidden representations of the RNN before feeding everything through a final linear layer.

\section{Experiments}
\subsection{Time Drift Experiments}
We seek to investigate how different models replicating commercial feature sets would age in clinical settings. To understand why model performance may change over time, we investigate technical and clinical sources of data shift.


\paragraph{Year Agnostic}
Similar to common model development practices, we report numbers from models trained and tested on the full set of MIMIC-IV patient data from 2008-2019, with patients randomly assigned to either training, validation or testing regardless of the year of care.

\paragraph{Year Buckets} 
To measure temporal model drift, we train on a subset of patient data in the 2008-2010 bucket, then test on patient data from the unseen portion of the 2008-2010 bucket as well as the entirety of the data in the subsequent three year buckets 2011-2013, 2014-2016, and 2017-2019. 


\subsection{Investigating Cause of Time Drift}
We investigate two causes of model drift: technical drift, and data drift. To investigate technical drift, we analyzed the effect of Beth Israel's switch in use of diagnoses codes from ICD-9 to ICD-10 in October 2015. We investigate the impact of these changing codes on model performance by removing all features from the Epic feature set that rely on ICD codes (HIV, obesity, coronary artery disease, congestive heart failure, chronic obstructive pulmonary disease, chronic kidney disease, chronic liver disease, diabetes, and hypertension), and retraining a new model with all other hyper-parameters kept constant. To investigate data shift over time, we plotted sepsis onset times, microbiology samples, and antibiotic usage trends over the years.


\section{Results}
We ran each combination of feature set, model, and training regime three times to measure the models' performance. The average AUCs from the three runs are reported in Tables \ref{epic_dascena_results} and \ref{epic_minus_icd_results}. Overall, we observe a large drop in performance for models trained on the Epic feature set, especially the RNN. The models trained on the Dascena model also experienced some performance degradation, but on a much smaller scale.

\subsection{Time Drift Experiments}
\begin{table}[H]
  \centering
  \begin{tabular}{llllll}
    \toprule
       &       & \multicolumn{4}{l}{Year Buckets}          \\
       \cmidrule(r){3-6}
 Model & Year-Agnostic & 2008-2010 & 2011-2013 & 2014-2016 & 2017-2019\\
    \midrule
\underline{Dascena Features} \\[0.2em]
RNN    & 0.764 & 0.723 & 0.726 & 0.730 & 0.707 \\
Logistic & 0.731 & 0.687 & 0.696 & 0.682 & 0.640 \\[0.5em]
\underline{Epic Features} \\[0.3em]
RNN    & 0.787 & 0.729 & 0.723 & 0.578 & 0.525 \\
Logistic & 0.724 & 0.706 & 0.701 & 0.679 & 0.626 \\
\bottomrule
  \end{tabular}
  \caption{Time agnostic and year bucket results (average test AUCs of three runs) from RNN and Logistic Regression models trained on the Epic and Dascena feature sets.}
  \label{epic_dascena_results}
\end{table}



We observe in \autoref{epic_dascena_results} that the models trained on Epic's feature set dropped from 0.729 to 0.525 for the RNN model, and from 0.706 to 0.626 for the Logistic Regression model. This amounts to a 0.08 to 0.20 difference in test AUC over the years. These results indicate that several Epic features vary over time. In contrast, models built on Dascena features, which contain mainly vital signs, perform significantly better over time. We found a small temporal drop in AUC from 0.723 to 0.707 for the RNN model and a slightly larger AUC drop from 0.687 to 0.640 for the Logistic Regression model. In addition, we note that the choice in model architecture impacts performance drop, as we found that there is a much larger drop for the RNN model than the Logistic Regression model trained on Epic's feature set, while the opposite was true for models trained on the Dascena feature set. We also include results for two additional time drift experiments, Length of Stay and ICU Mortality prediction, in Appendix B. 

\subsection{Time Drift Investigations}
\paragraph{Technical Change Investigation}

\begin{table}[H]
  \centering
  \begin{tabular}{llllll}
    \toprule
       &       & \multicolumn{4}{l}{Year Buckets}          \\
       \cmidrule(r){3-6}
Model & Year-Agnostic & 2008-2010 & 2011-2013 & 2014-2016 & 2017-2019\\
    \midrule
RNN    &  0.783     & 0.716 & 0.731 & 0.721 & 0.671 \\
Logistic &  0.669     & 0.715 & 0.702 & 0.692 & 0.633 \\
    \bottomrule
  \end{tabular}
  \caption{Average test AUCs from models trained on the Epic feature subset without ICD codes.}
  \label{epic_minus_icd_results}
\end{table}


In \autoref{epic_minus_icd_results}, we report the results of models trained on a subset of the Epic feature set without ICD codes. We found that without the ICD codes, the AUC drop across the entire time period is now 0.08 for the Logistic Regression model and 0.045 for the RNN. Compared to the RNN model trained on the full Epic feature set, this represents a significant improvement of around 0.15 for the last year bucket. Because models trained on the first year bucket only observed ICD-9 codes, they were unable to leverage ICD-10 features when the transition occurred. 



We notice that the introduction of the ICD codes did not contribute to higher overall model performance, as seen by the year-agnostic numbers in \autoref{epic_dascena_results} and \autoref{epic_minus_icd_results}. However, the inclusion of these features instead led to significant model degradation.

\paragraph{Data Change Investigation}

\begin{figure}[H]
\centering
\includegraphics[width=0.3\textwidth]{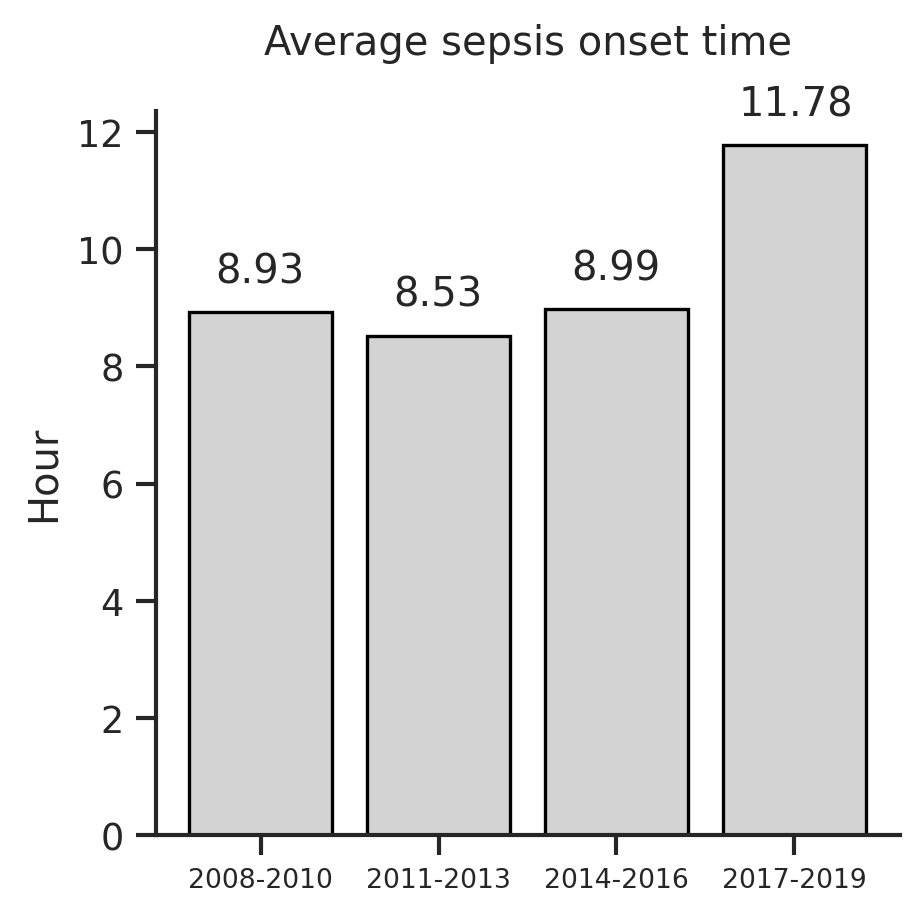}
\caption{Sepsis onset time in the ICU by year bucket groups.}
\label{fig:sepsis_onset}
\end{figure}

\begin{figure}[H]
\centering
\includegraphics[width=0.50\textwidth]{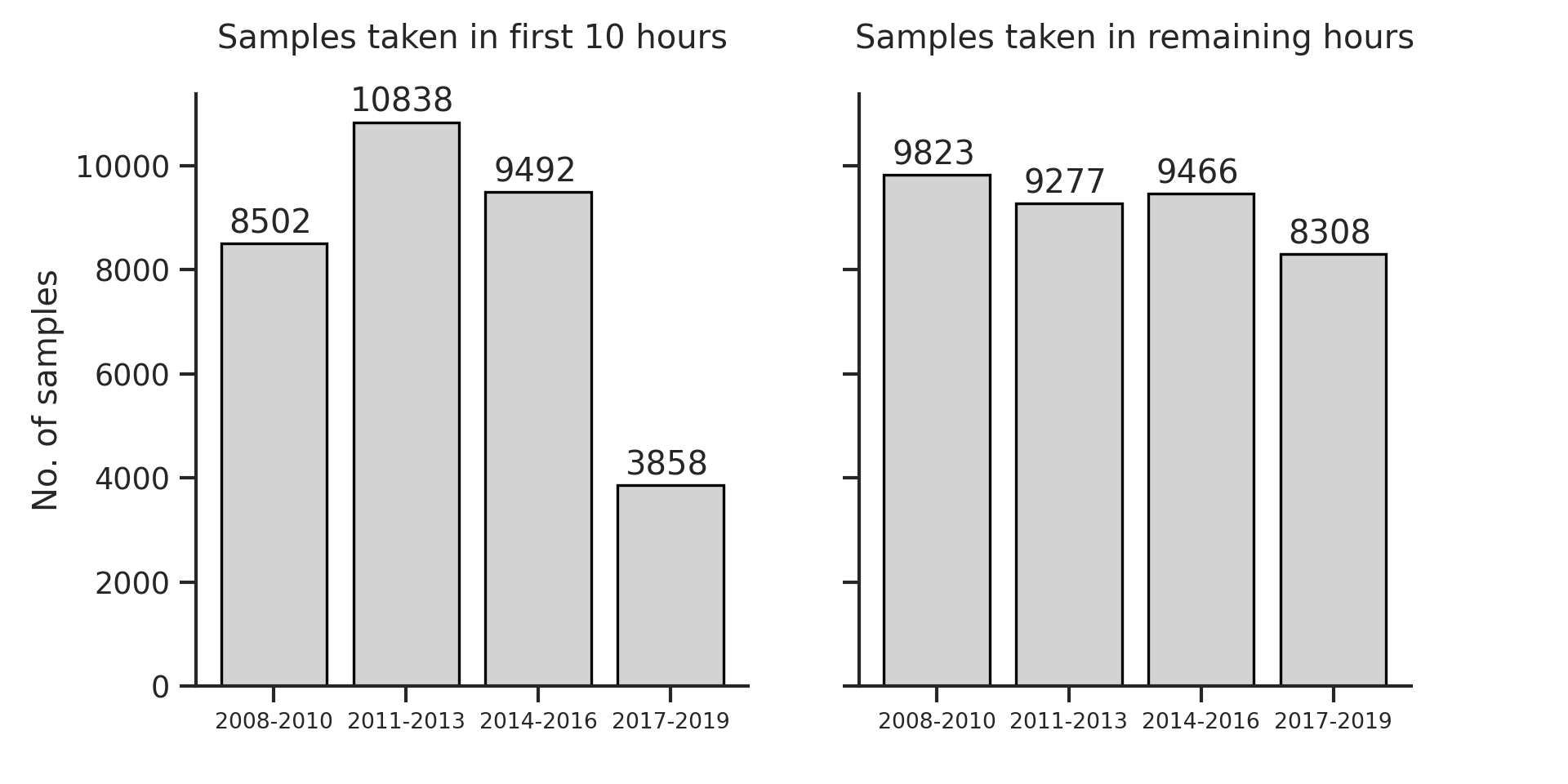}
\caption{Microbiology samples across year buckets and time into ICU stay.}
\label{fig:cultures_changing}
\end{figure}

\begin{figure}[H]
\centering
\includegraphics[width=0.50\textwidth]{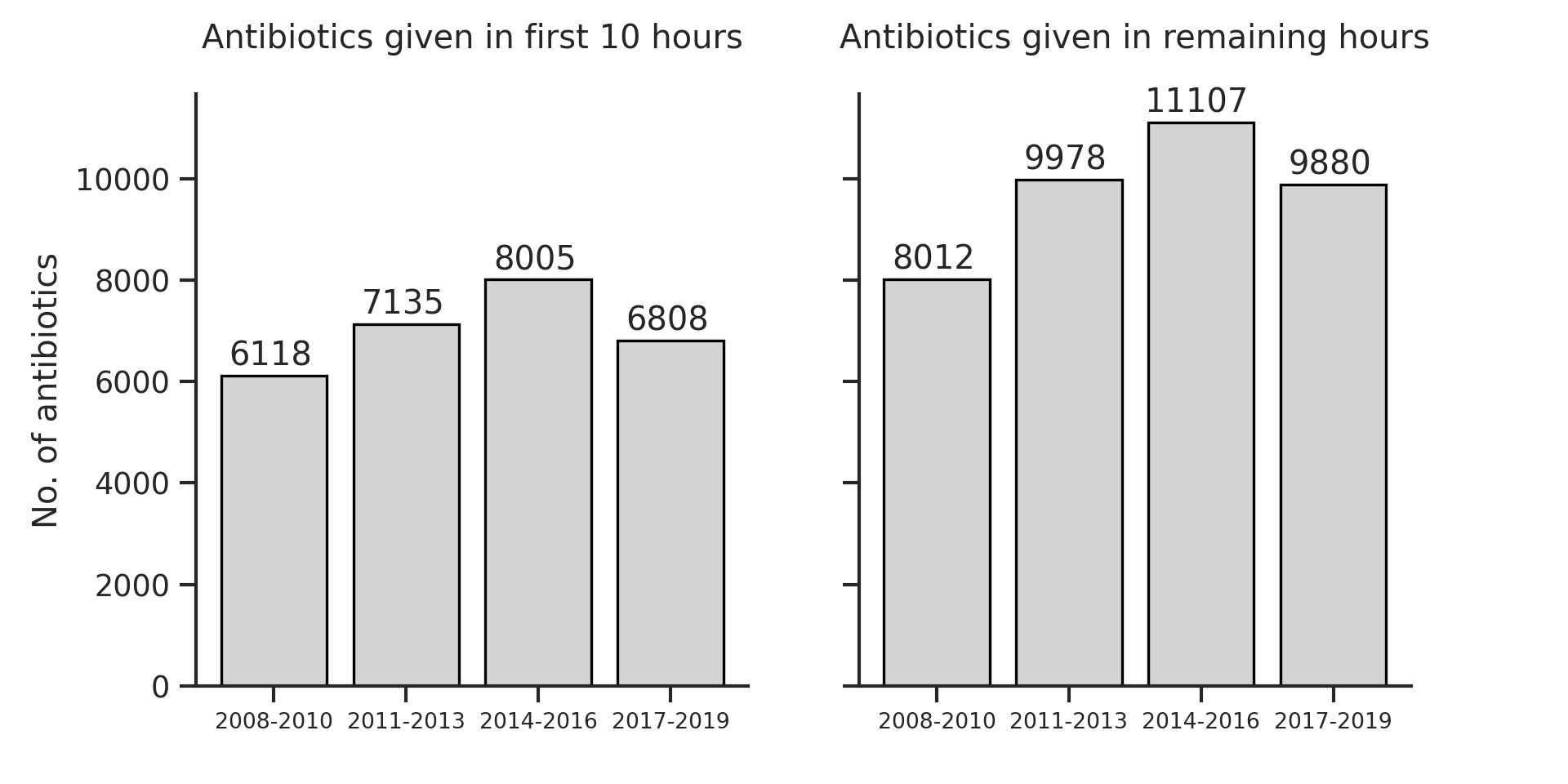}
\caption{Antibiotics administered across year buckets and time into ICU stay.}
\label{fig:antibio_changing}
\end{figure}

\begin{figure}[H]
\centering
\includegraphics[width=0.95\textwidth]{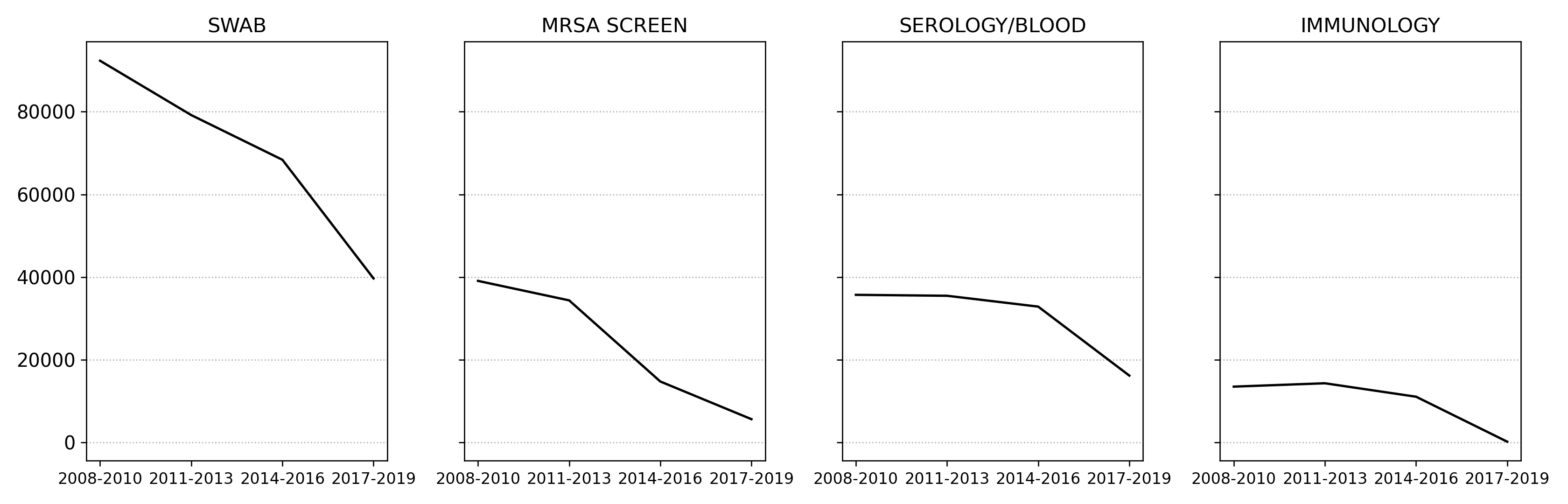}
\caption{Microbiological culture types with the largest drop in number of samples taken across year bucket groups. }
\label{fig:select_cultures}
\end{figure}

\begin{table}[H]
\begin{tabular}{rrrrrrr}
\toprule
\textbf{Specimen Type}              & \textbf{2008-2010} & \textbf{2011-2013} & \textbf{2014-2016} & \textbf{2017-2019} & \textbf{Change} & \textbf{\% Change} \\ \midrule
\scriptsize
\textbf{SWAB}                       & 92304              & 79169              & 68363              & 39677              & -52627          & -57\%                   \\
 
\scriptsize\textbf{MRSA SCREEN}                & 39086              & 34375              & 14766              & 5657               & -33429          & -86\%                   \\
 
\scriptsize\textbf{BLOOD CULTURE}              & 171630             & 158924             & 163711             & 144620             & -27010          & -16\%                   \\
 
\scriptsize\textbf{SPUTUM}                     & 52437              & 39027              & 40078              & 32135              & -20302          & -39\%                   \\
 
\scriptsize\textbf{SEROLOGY/BLOOD}             & 35713              & 35489              & 32866              & 16187              & -19526          & -55\%                   \\
 
\scriptsize\textbf{STOOL}                      & 47316              & 38649              & 38166              & 28502              & -18814          & -40\%                   \\
 
\scriptsize\textbf{IMMUNOLOGY}                 & 13524              & 14336              & 11085              & 209                & -13315          & -98\%                   \\
 
\scriptsize\textbf{URINE}                      & 221335             & 211626             & 263363             & 234576             & +13241           & +6\%                    \\
 
\textbf{\scriptsize{}BONE MARROW - CYTOGENETICS} & 7442               & 3643               & 5                  & 0                  & -7442           & -100\%                  \\
 
\scriptsize\textbf{TISSUE}                     & 21455              & 26311              & 32957              & 28481              & +7026            & +33\%                   \\
 
\scriptsize\textbf{Immunology (CMV)}           & 5921               & 5278               & 4285               & 42                 & -5879           & -99\%                   \\
 
\scriptsize\textbf{CATHETER OR LINE}           & 6862               & 3653               & 2014               & 1397               & -5465           & -80\%                   \\
 
\scriptsize\textbf{CSF;SPINAL FLUID}           & 10725              & 8527               & 8601               & 6333               & -4392           & -41\%                   \\
 
\scriptsize\textbf{BRONCHIAL WASHINGS}         & 2214               & 2577               & 5087               & 6442               & +4228            & +191\%                  \\
 
\scriptsize\textbf{Influenza A/B by DFA}       & 4214               & 2771               & 532                & 0                  & -4214           & -100\%                  \\
 
\scriptsize\textbf{ABSCESS}                    & 9597               & 11657              & 12734              & 13796              & +4199            & +44\%                   \\
 
\scriptsize\textbf{Blood (LYME)}               & 0                  & 4                  & 2414               & 3815               & +3815            & +100\%                  \\
 
\scriptsize\textbf{PLEURAL FLUID}              & 7560               & 7560               & 8834               & 9826               & +2266            & -30\%                   \\
 
\scriptsize\textbf{BRONCHOALVEOLAR LAVAGE}     & 12772              & 10930              & 13205              & 10523              & -2249           & -18\%                   \\
\scriptsize\textbf{Staph aureus Screen}        & 3609               & 9151               & 11949              & 5608               & 1999            & +55\% \\\bottomrule
\end{tabular}
\caption{Specimen types with greatest absolute change between 2008-2010 and 2017-2019 year buckets. ``Change'' and ``\% Change'' refers to the difference between the first and the last year buckets.}\label{tab:spec_changes}
\end{table}

\begin{figure}[H]
\centering
\includegraphics[width=0.9\textwidth]{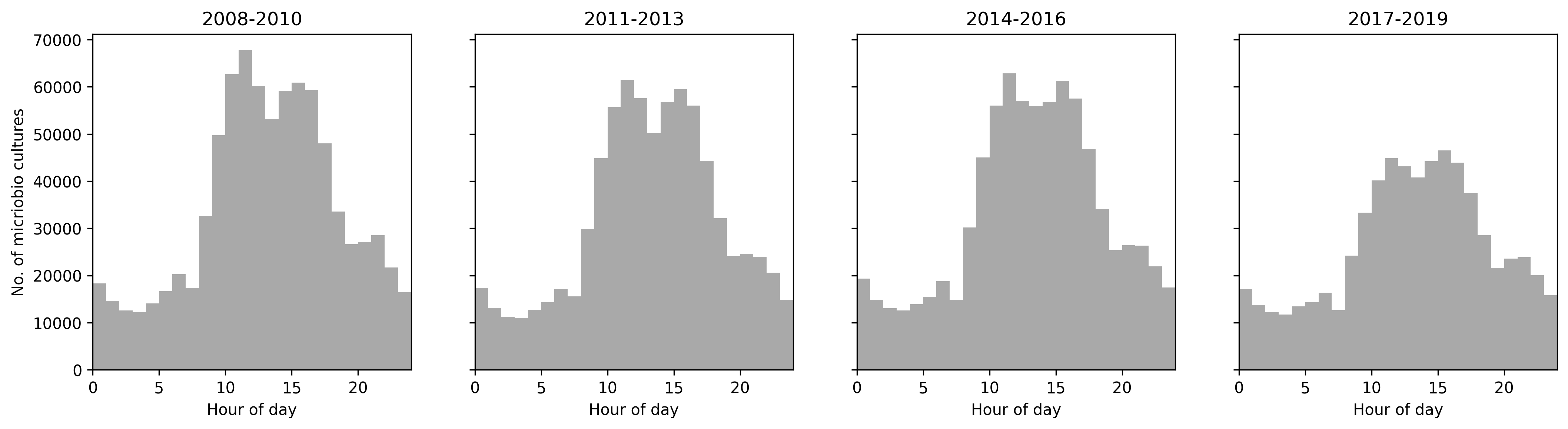}
\caption{Amount of cultures drawn for each hour of the day, per year-bucket.}
\label{fig:microbio_day}
\end{figure}


We plot all data changes in Figures \ref{fig:sepsis_onset}-\ref{fig:microbio_day} and Table \ref{tab:spec_changes}. While we see no major changes in antibiotics administered, we see a large increase in average sepsis onset time from 9 hours to 12 hours by the last year bucket, accompanied by a drop in microbiology samples taken within the first 10 hours of a patient's ICU stay. \autoref{fig:select_cultures} shows a select few of the microbiological sample types with the largest changes across the years, and \autoref{tab:spec_changes} shows the full numbers in detail. Since the definition of Sepsis-3 onset depends on a suspicion of infection, which is influenced by the timing of microbiology samples, we observe that these two changes are related. Finally, \autoref{fig:microbio_day} shows that the drop in number of samples occurs mainly during daytime hours (between 9 AM and 5 PM).

Clinical sources we interviewed explained that changing hospital demographics, caused by multiple new hospital acquisitions and partnerships, in recent years could have been the cause of changing clinical practices around microbiology sampling and other ICU procedures.



\section{Conclusion}
By replicating commercially produced sepsis prediction models on the MIMIC-IV dataset, we were able to observe the extent of model degradation over time. We found a significant AUC drop of 0.08 to 0.20 for models trained on the Epic feature set, while a smaller drop of 0.02 to 0.05 was observed for models trained on the more stable Dascena feature set. We concluded that the technical shift of changing ICD codes from ICD-9 to ICD-10 in 2015 was a major cause of degradation in the former models, as the performance drop diminished by 0.15 after the models were retrained without ICD codes. Finally, we observed a data shift in later years from changing microbiology sampling practices in the ICU, which had an impact on delaying sepsis onset times in the last year bucket. 

\paragraph{Limitations}
There are several limitations to this study. First, because the Epic and Dascena models are not publicly available, we trained our models leveraging their published feature sets on MIMIC-IV. These features can be found in Appendix A. Our results do not directly reflect the performance of Epic or Dascena's commercial models. Second, this study does not account for possible model retraining procedures a hospital could undertake to diminish the impact of model degradation in clinical settings.

\section*{Acknowledgements}
We are grateful to BIDMC, the MIMIC team, Leo Anthony Celi and Brian Gow for their support of this project.

\section*{Data and Code Availability}
The MIMIC-IV dataset is freely available provided you have completed CITI certification. It can be downloaded from \url{https://physionet.org/content/mimiciv/1.0/}.

All code used to reproduce the results in this report (including instructions on how to do so) is openly available at \url{https://github.com/mariehane/ai-gone-astray}.

\section*{References}
{
\small
\printbibliography[heading=none]

@misc{johnson_alistair_mimic-iv_nodate,
	title = {{MIMIC}-{IV}},
	url = {https://physionet.org/content/mimiciv/0.4/},
	urldate = {2022-02-19},
	publisher = {PhysioNet},
	author = {Johnson, Alistair and Bulgarelli, Lucas and Pollard, Tom and Horng, Steven and Celi, Leo Anthony and Mark, Roger},
	doi = {10.13026/A3WN-HQ05},
	note = {Version Number: 0.4
Type: dataset},
}

@article{nestor_feature_2019,
	title = {Feature {Robustness} in {Non}-stationary {Health} {Records}: {Caveats} to {Deployable} {Model} {Performance} in {Common} {Clinical} {Machine} {Learning} {Tasks}},
	shorttitle = {Feature {Robustness} in {Non}-stationary {Health} {Records}},
	url = {http://arxiv.org/abs/1908.00690},
	urldate = {2022-02-19},
	journal = {arXiv:1908.00690 [cs, stat]},
	author = {Nestor, Bret and McDermott, Matthew B. A. and Boag, Willie and Berner, Gabriela and Naumann, Tristan and Hughes, Michael C. and Goldenberg, Anna and Ghassemi, Marzyeh},
	month = aug,
	year = {2019},
	note = {arXiv: 1908.00690},
}

@article{moor_predicting_2021,
	title = {Predicting sepsis in multi-site, multi-national intensive care cohorts using deep learning},
	url = {http://arxiv.org/abs/2107.05230},
	urldate = {2022-02-19},
	journal = {arXiv:2107.05230 [cs]},
	author = {Moor, Michael and Bennet, Nicolas and Plecko, Drago and Horn, Max and Rieck, Bastian and Meinshausen, Nicolai and Bühlmann, Peter and Borgwardt, Karsten},
	month = jul,
	year = {2021},
	note = {arXiv: 2107.05230},
}

@article{che_recurrent_2018,
	title = {Recurrent {Neural} {Networks} for {Multivariate} {Time} {Series} with {Missing} {Values}},
	volume = {8},
	copyright = {2018 The Author(s)},
	issn = {2045-2322},
	url = {https://www.nature.com/articles/s41598-018-24271-9},
	doi = {10.1038/s41598-018-24271-9},
	language = {en},
	number = {1},
	urldate = {2022-02-19},
	journal = {Scientific Reports},
	author = {Che, Zhengping and Purushotham, Sanjay and Cho, Kyunghyun and Sontag, David and Liu, Yan},
	month = apr,
	year = {2018},
	note = {Number: 1
Publisher: Nature Publishing Group},
	pages = {6085},
}

@inproceedings{moor_early_2019,
	title = {Early {Recognition} of {Sepsis} with {Gaussian} {Process} {Temporal} {Convolutional} {Networks} and {Dynamic} {Time} {Warping}},
	url = {https://proceedings.mlr.press/v106/moor19a.html},
	language = {en},
	urldate = {2022-02-19},
	booktitle = {Proceedings of the 4th {Machine} {Learning} for {Healthcare} {Conference}},
	publisher = {PMLR},
	author = {Moor, Michael and Horn, Max and Rieck, Bastian and Roqueiro, Damian and Borgwardt, Karsten},
	month = oct,
	year = {2019},
	note = {ISSN: 2640-3498},
	pages = {2--26},
}

@article{barton_evaluation_2019,
	title = {Evaluation of a machine learning algorithm for up to 48-hour advance prediction of sepsis using six vital signs},
	volume = {109},
	issn = {0010-4825},
	url = {https://www.ncbi.nlm.nih.gov/pmc/articles/PMC6556419/},
	doi = {10.1016/j.compbiomed.2019.04.027},
	urldate = {2022-02-20},
	journal = {Computers in biology and medicine},
	author = {Barton, Christopher and Chettipally, Uli and Zhou, Yifan and Jiang, Zirui and Lynn-Palevsky, Anna and Le, Sidney and Calvert, Jacob and Das, Ritankar},
	month = jun,
	year = {2019},
	pmid = {31035074},
	pmcid = {PMC6556419},
	pages = {79--84},
}

@article{elman_finding_1990,
	title = {Finding {Structure} in {Time}},
	volume = {14},
	issn = {1551-6709},
	url = {https://onlinelibrary.wiley.com/doi/abs/10.1207/s15516709cog1402_1},
	doi = {10.1207/s15516709cog1402_1},
	language = {en},
	number = {2},
	urldate = {2022-02-24},
	journal = {Cognitive Science},
	author = {Elman, Jeffrey L.},
	year = {1990},
	note = {\_eprint: https://onlinelibrary.wiley.com/doi/pdf/10.1207/s15516709cog1402\_1},
	pages = {179--211},
}
}

\clearpage
\appendix

\raggedbottom 
\section{Feature Lists and Origin Categories}
Below are the features we found in MIMIC-IV that best approximate the feature categories used by Dascena and Epic's models, given that their code is not publicly available. For Epic's feature set, we were unable to find certain feature results in the MIMIC database, including medicines administered by category and the exact mapping of ICD codes for all diagnoses used. 

\subsection{Dascena Features}
\begin{table}[H]
  \centering
  \tiny
  \begin{tabular}{llll}
  \toprule
\textbf{High-level feature}                           & \textbf{Itemid}          & \textbf{Low-level label}                 & \textbf{Origin}          \\\midrule
\multirow{8}{*}{diastolic blood pressure} & 8441                       & nbp {[}diastolic{]}                   & chartevents                \\
                                          & 8368                       & arterial bp {[}diastolic{]}           & chartevents                \\
                                          & 220180                     & non invasive blood pressure diastolic & chartevents                \\
                                          & 220051                     & arterial blood pressure diastolic     & chartevents                \\
                                          & 225310                     & art bp diastolic                      & chartevents                \\
                                          & 8555                       & arterial bp \#2 {[}diastolic{]}       & chartevents                \\
                                          & 8440                       & manual bp {[}diastolic{]}             & chartevents                \\
                                          & 224643                     & manual blood pressure diastolic left  & chartevents                \\\midrule
\multirow{9}{*}{systolic blood pressure}  & 455                        & nbp {[}systolic{]}                    & chartevents                \\
                                          & 51                         & arterial bp {[}systolic{]}            & chartevents                \\
                                          & 220179                     & non invasive blood pressure systolic  & chartevents                \\
                                          & 220050                     & arterial blood pressure systolic      & chartevents                \\
                                          & 225309                     & art bp systolic                       & chartevents                \\
                                          & 6701                       & arterial bp \#2 {[}systolic{]}        & chartevents                \\
                                          & 442                        & manual bp {[}systolic{]}              & chartevents                \\
                                          & 224167                     & manual blood pressure systolic left   & chartevents                \\
                                          & 227243                     & manual blood pressure systolic right  & chartevents                \\\midrule
\multirow{2}{*}{heart rate}               & 211                        & heart rate                            & chartevents                \\
                                          & 220045                     & heart rate                            & chartevents                \\\midrule
\multirow{6}{*}{temperature}              & 678                        & temperature f                         & chartevents                \\
                                          & 677                        & temperature c (calc)                  & chartevents                \\
                                          & 223761                     & temperature fahrenheit                & chartevents                \\
                                          & 679                        & temperature f (calc)                  & chartevents                \\
                                          & 676                        & temperature c                         & chartevents                \\
                                          & 223762                     & temperature celsius                   & chartevents                \\\midrule
\multirow{8}{*}{respiratory rate}         & 618                        & respiratory rate                      & chartevents                \\
                                          & 220210                     & respiratory rate                      & chartevents                \\
                                          & 615                        & resp rate (total)                     & chartevents                \\
                                          & 614                        & resp rate (spont)                     & chartevents                \\
                                          & 224689                     & respiratory rate (spontaneous)        & chartevents                \\
                                          & 224690                     & respiratory rate (total)              & chartevents                \\
                                          & 651                        & spon rr (mech.)                       & chartevents                \\
                                          & 224422                     & spont rr                              & chartevents                \\\midrule
\multirow{4}{*}{oxygen saturation}        & 646                        & spo2                                  & chartevents                \\
                                          & 220277                     & o2 saturation pulseoxymetry           & chartevents                \\
                                          & 834                        & sao2                                  & chartevents                \\
                                          & 220227                     & arterial o2 saturation                & chartevents               \\
  \bottomrule
\end{tabular}
\caption{Full Dascena feature set}
\end{table}

\subsection{Epic Features}\label{sec:epic_features}

\begin{table}[H]
  \centering
  \tiny
  \begin{tabular}{ll}
  \toprule
  \textbf{Feature} & \textbf{Origin} \\
  \midrule
age              & patients                            \\ \midrule
ethnicity        & admissions                          \\ \midrule
martial status   & admissions                          \\ \midrule
gender           & patients \\ \bottomrule
\end{tabular}
\caption{Static demographical features used by Epic's model}
\end{table}

\begin{table}[H]
  \centering
  \tiny
  \begin{tabular}{llll}
  \toprule
\textbf{High-level feature}                           & \textbf{Itemid}          & \textbf{Low-level label}                 & \textbf{Origin}          \\\midrule
\multirow{3}{*}{creatinine}                  & 791                                 & creatinine (0-1.3)              & chartevents     \\

                                             & 1525                                & creatinine                      & chartevents     \\
                                             & 220615                              & creatinine                      & chartevents     \\ \midrule
\multirow{2}{*}{heart rate}                  & 211                                 & heart rate                      & chartevents     \\
                                             & 220045                              & heart rate                      & chartevents     \\ \midrule
\multirow{2}{*}{hematocrit}                  & 813                                 & hematocrit                      & chartevents     \\
                                             & 220545                              & hematocrit (serum)              & chartevents     \\ \midrule
\multirow{2}{*}{hemoglobin}                  & 814                                 & hemoglobin                      & chartevents     \\
                                             & 220228                              & hemoglobin                      & chartevents     \\ \midrule
\multirow{2}{*}{platelets}                   & 828                                 & platelets                       & chartevents     \\
                                             & 227457                              & platelet count                  & chartevents     \\ \midrule
red blood cell count                         & 833                                 & rbc                             & chartevents     \\ \midrule
\multirow{8}{*}{respiratory rate}            & 614                                 & resp rate (spont)               & chartevents     \\
                                             & 615                                 & resp rate (total)               & chartevents     \\
                                             & 618                                 & respiratory rate                & chartevents     \\
                                             & 651                                 & spon rr (mech.)                 & chartevents     \\
                                             & 220210                              & respiratory rate                & chartevents     \\
                                             & 224422                              & spont rr                        & chartevents     \\
                                             & 224689                              & respiratory rate (spontaneous)  & chartevents     \\
                                             & 224690                              & respiratory rate (total)        & chartevents     \\ \midrule
\multirow{2}{*}{respiratory rate set}        & 619                                 & respiratory rate set            & chartevents     \\
                                             & 224688                              & respiratory rate (set)          & chartevents     \\ \midrule
\multirow{6}{*}{temperature}                 & 676                                 & temperature c                   & chartevents     \\
                                             & 677                                 & temperature c (calc)            & chartevents     \\
                                             & 678                                 & temperature f                   & chartevents     \\
                                             & 679                                 & temperature f (calc)            & chartevents     \\
                                             & 223761                              & temperature fahrenheit          & chartevents     \\
                                             & 223762                              & temperature celsius             & chartevents     \\ \midrule
\multirow{4}{*}{white blood cell count}      & 861                                 & wbc (4-11,000)                  & chartevents     \\
                                             & 1127                                & wbc (4-11,000)                  & chartevents     \\
                                             & 1542                                & wbc                             & chartevents     \\
                                             & 220546                              & wbc                             & chartevents     \\ \midrule
Band neutrophils                             & 51144                               & Bands                           & labevents       \\ \midrule
Base excess                                  & 50802                               & Base Excess                     & labevents       \\ \midrule
Lymphocyte                                   & 51244                               & Lymphocytes                     & labevents       \\ \midrule
Mean corpuscular hemoglobin concentration    & 51249                               & MCHC                            & labevents       \\ \midrule
Monocytes                                    & 51254                               & Monocytes                       & labevents       \\ \midrule
Neutrophils                                  & 51256                               & Neutrophils                     & labevents       \\ \midrule
Nucleated red blood cell count               & 51257                               & Nucleated Red Cells             & labevents       \\ \midrule
Red Blood Cell morphology                    & 52171                               & RBC Morphology                  & labevents       \\ \midrule
Red Blood Cell distribution width            & 52204                               & RBCDist                         & labevents       \\ \midrule
reticulocyte count                           & 51282                               & Reticulocyte Count, Absolute    & labevents       \\ \midrule
Segmented neutrophil count                   & 51232                               & Hypersegmented Neutrophils      & labevents       \\ \bottomrule
\end{tabular}
\end{table}

\begin{table}[H]
  \centering
  \tiny
  \begin{tabular}{llll}
  \toprule
\textbf{High-level feature}                           & \textbf{Itemid}          & \textbf{Low-level label}                 & \textbf{Origin}          \\\midrule
Peripherally inserted central catheters      & 224264                              & PICC Line                       & procedureevents \\ \midrule
Central venous catheters                     & 225315                              & Tunneled (Hickman) Line         & procedureevents \\ \midrule
\multirow{5}{*}{Drains}                      & 225447                              & percutaneous drain insertion    & procedureevents \\
                                             & 225456                              & ventricular drain               & procedureevents \\
                                             & 226475                              & intraventricular drain inserted & procedureevents \\
                                             & 229523                              & subdural drain                  & procedureevents \\
                                             & 229524                              & lumbar drain                    & procedureevents \\ \midrule
\multirow{3}{*}{Feeding tube}                & 224007                              & gi \#1 intub site               & chartevents     \\
                                             & 224441                              & gi \#2 intub site               & chartevents     \\
                                             & 224442                              & gi \#3 intub site               & chartevents     \\ \midrule
\multirow{16}{*}{Incision}                   & 227472                              & incision site \#1               & chartevents     \\
                                             & 227473                              & incision site \#2               & chartevents     \\
                                             & 227474                              & incision site \#3               & chartevents     \\
                                             & 227475                              & incision site \#4               & chartevents     \\
                                             & 227476                              & incision site \#5               & chartevents     \\
                                             & 227477                              & incision site \#6               & chartevents     \\
                                             & 228559                              & incision \#1- location          & chartevents     \\
                                             & 228560                              & incision \#2- location          & chartevents     \\
                                             & 228561                              & incision \#3- location          & chartevents     \\
                                             & 228562                              & incision \#4- location          & chartevents     \\
                                             & 228563                              & incision \#5- location          & chartevents     \\
                                             & 228564                              & incision \#6- location          & chartevents     \\
                                             & 229015                              & incision \#7- location          & chartevents     \\
                                             & 229016                              & incision \#8- location          & chartevents     \\
                                             & 229017                              & incision \#9- location          & chartevents     \\
                                             & 229018                              & incision \#10- location         & chartevents     \\ \midrule
\multirow{10}{*}{Pressure Ulcers}            & 228506                              & pressure ulcer \#1- location    & chartevents     \\
                                             & 228507                              & pressure ulcer \#2- location    & chartevents     \\
                                             & 228508                              & pressure ulcer \#3- location    & chartevents     \\
                                             & 228509                              & pressure ulcer \#4- location    & chartevents     \\
                                             & 228510                              & pressure ulcer \#5- location    & chartevents     \\
                                             & 228511                              & pressure ulcer \#6- location    & chartevents     \\
                                             & 228512                              & pressure ulcer \#7- location    & chartevents     \\
                                             & 228513                              & pressure ulcer \#8- location    & chartevents     \\
                                             & 228514                              & pressure ulcer \#9- location    & chartevents     \\
                                             & 228515                              & pressure ulcer \#10- location   & chartevents     \\ \midrule
\multirow{2}{*}{Active penicillin orders}    & 008880                              & Penicillin V Potassium          & prescriptions   \\
                                             & 043350                              & Penicillin G Benzathine         & prescriptions   \\ \midrule
\multirow{6}{*}{Active vancomycin orders}    & 043952                              & Vancomycin                      & prescriptions   \\
                                             & 009331                              & Vancomycin                      & prescriptions   \\
                                             & 009328                              & Vancomycin                      & prescriptions   \\
                                             & 009329                              & Vancomycin                      & prescriptions   \\
                                             & 067111                              & Vancomycin                      & prescriptions   \\
                                             & 020611                              & Vancomycin                      & prescriptions \\
\bottomrule
 \end{tabular}
  \caption{Time-varying Epic features.}
\end{table}

\begin{table}[H]
  \centering
  \tiny
  \begin{tabular}{llll}
  \toprule
\textbf{Feature}                           & \textbf{ICD Code}          & \textbf{ICD Version}                 & \textbf{Origin}          \\\midrule
\multirow{2}{*}{Diabetes}                    & \multicolumn{1}{l}{E11}             & ICD-10                          & diagnoses\_icd  \\
                                             & \multicolumn{1}{l}{E10}             & ICD-10                          & diagnoses\_icd  \\ \midrule
Hypertension                                 & \multicolumn{1}{l}{I10}             & ICD-10                          & diagnoses\_icd  \\ \midrule
\multirow{2}{*}{HIV}                         & 42                                  & ICD-9                           & diagnoses\_icd  \\
                                             & \multicolumn{1}{l}{B20}             & ICD-10                          & diagnoses\_icd  \\ \midrule
\multirow{2}{*}{Obesity}                     & 27800                               & ICD-9                           & diagnoses\_icd  \\
                                             & \multicolumn{1}{l}{E66}             & ICD-10                          & diagnoses\_icd  \\ \midrule
\multirow{2}{*}{Coronary Artery Disease}     & 41400                               & ICD-9                           & diagnoses\_icd  \\
                                             & \multicolumn{1}{l}{I251}            & ICD-10                          & diagnoses\_icd  \\ \midrule
\multirow{3}{*}{Congestive Heart Failure}    & \multicolumn{1}{l}{I502}            & ICD-10                          & diagnoses\_icd  \\
                                             & \multicolumn{1}{l}{I503}            & ICD-10                          & diagnoses\_icd  \\
                                             & \multicolumn{1}{l}{I504}            & ICD-10                          & diagnoses\_icd  \\ \midrule
Chronic Obstructive Pulmonary Disease (COPD) & \multicolumn{1}{l}{J44}             & ICD-10                          & diagnoses\_icd  \\ \midrule
\multirow{2}{*}{Chronic Kidney Disease}      & \multicolumn{1}{l}{I13}             & ICD-10                          & diagnoses\_icd  \\
                                             & \multicolumn{1}{l}{I12}             & ICD-10                          & diagnoses\_icd  \\ \midrule
\multirow{2}{*}{Chronic Liver Disease}       & 5719                                & ICD-9                           & diagnoses\_icd  \\
                                             & \multicolumn{1}{l}{K76}             & ICD-10                          & diagnoses\_icd  \\
\bottomrule
 \end{tabular}
  \caption{Epic features based on ICD-codes.}
\end{table}

\section{Other Prediction Tasks}
To validate our results of temporal drift with Nestor et al.'s \cite{nestor_feature_2019} experiments on MIMIC-III, we followed a similar setup to run experiments on Length of Stay and ICU Mortality prediction tasks on MIMIC-IV, with results in \autoref{los_raw_results} and \autoref{icu_raw_results}. The length of stay prediction task involves taking the first 24 hours of data of a patient’s stay in the ICU, and predicting whether this patient’s specific stay in the ICU would last >= 3 days (a binary prediction task). The ICU mortality task uses the same feature set as the LOS prediction task, and likewise uses the first 24 hours of data of a patient’s ICU stay to predict whether a patient would die in the ICU during this particular stay. The features used for both prediction tasks were the same 8 demographic variables and 141 vital signs from the chartevents table used by Nestor et al. 


\subsection{Length-of-stay (LOS) Results}
\begin{table}[H]
  \centering
  \begin{tabular}{llllll}
    \toprule
       &       & \multicolumn{4}{l}{Year Buckets}          \\
       \cmidrule(r){3-6}
Model & Year-Agnostic & 2008-2010 & 2011-2013 & 2014-2016 & 2017-2019\\
    \midrule
RNN    & 0.658 & 0.661 & 0.635 & 0.623 & 0.564 \\
Logistic & 0.601 & 0.620 & 0.600 & 0.585 & 0.549 \\
\bottomrule
  \end{tabular}
  \caption{Results from Length-of-stay prediction on MIMIC-IV, trained on raw data representations}
  \label{los_raw_results}
\end{table}


\subsection{ICU Mortality Results}

\begin{table}[H]
  \centering
  \begin{tabular}{llllll}
    \toprule
       &       & \multicolumn{4}{l}{Year Buckets}          \\
       \cmidrule(r){3-6}
Model & Year-Agnostic & 2008-2010 & 2011-2013 & 2014-2016 & 2017-2019\\
    \midrule
RNN    & 0.830 & 0.745 & 0.760 & 0.776 & 0.777 \\
Logistic & 0.680 & 0.696 & 0.684 & 0.693 & 0.690 \\
\bottomrule
  \end{tabular}
  \caption{Results from ICU Mortality prediction on MIMIC-IV, trained on raw data representations}
  \label{icu_raw_results}
\end{table}


\end{document}